\documentclass[sigconf]{acmart}

\usepackage{booktabs} 

\usepackage[ruled]{algorithm2e}
\usepackage{flushend}
\setcopyright{rightsretained}

\acmDOI{}

\acmISBN{}

\acmConference[SIGIR 2018 eCom]{ACM SIGIR Workshop on eCommerce}{July 2018}{Ann Arbor, Michigan, USA}
\acmYear{2018}
\copyrightyear{2018}

\acmArticle{}
\acmPrice{}

\begin{document}
\title{Speeding up the Metabolism in E-commerce by Reinforcement Mechanism Design }

\author{Hua-Lin He}
\orcid{0000-0001-6441-7700}
\affiliation{%
  \institution{Alibaba Inc.}
  \streetaddress{Zone Xixi, No. 969, West Wenyi Road}
  \city{Hangzhou}
  \state{China}
  \postcode{310000}
}
\email{hualin.hhl@alibaba-inc.com}

\author{Chun-Xiang Pan}
\affiliation{%
  \institution{Alibaba Inc.}
  \streetaddress{Zone Xixi, No. 969, West Wenyi Road}
  \city{Hangzhou}
  \state{China}
  \postcode{310000}
}
\email{xuanran@taobao.com}

\author{Qing Da}
\affiliation{%
  \institution{Alibaba Inc.}
  \streetaddress{Zone Xixi, No. 969, West Wenyi Road}
  \city{Hangzhou}
  \state{China}
  \postcode{310000}
 }
\email{daqing.dq@alibaba-inc.com}

\author{An-Xiang Zeng}
\affiliation{%
  \institution{Alibaba Inc.}
  \streetaddress{Zone Xixi, No. 969, West Wenyi Road}
  \city{Hangzhou}
  \state{China}
  \postcode{310000}
}
 \email{renzhong@taobao.com}

\renewcommand{\shortauthors}{Hua-Lin He et al.}

\begin{abstract}
In a large E-commerce platform, all the participants compete for impressions under the allocation mechanism of the platform. Existing methods mainly focus on the short-term return based on the current observations instead of the long-term return. In this paper, we formally establish the lifecycle model for products, by defining the \emph{introduction}, \emph{growth}, \emph{maturity} and \emph{decline} stages and their transitions throughout the whole life period. Based on such  model, we further propose a reinforcement learning based mechanism design framework for impression allocation, which incorporates the first principal component based permutation and the novel experiences generation method, to maximize short-term as well as long-term return of the platform. With the power of trial-and-error, it is possible to optimize impression allocation strategies globally which is contribute to the healthy development of participants and the platform itself. We evaluate our algorithm on a simulated environment built based on one of the largest E-commerce platforms, and a significant improvement has been achieved in comparison with the baseline solutions.
\end{abstract}

\ccsdesc[300]{Computing methodologies~Reinforcement learning}
\ccsdesc[300]{Computing methodologies~Policy iteration}
\ccsdesc[300]{Applied computing~Online shopping}

\keywords{Reinforcement Learning, Mechanism Design, E-commerce}

\maketitle

\section{Introduction}
Nowadays, E-commerce platform like Amazon or Taobao has developed into a large business ecosystem consisting of millions of customers, enterprises and start-ups, and hundreds of thousands of service providers, making it a new type of economic entity rather than enterprise platform. In such a economic entity, a major responsibility of the platform is to design economic institutions to achieve various business goals, which is the exact field of \emph{Mechanism Design}~\cite{vickrey1961counterspeculation}. Among all the affairs of the E-commerce platform, impression allocation is one of the key strategies to achieve its business goal, while products are players competing for the resources under the allocation mechanism of the platform, and the platform is the game designer aiming to design game whose outcome will be as the platform desires.

Existing work of impression allocation in literature are mainly motivated and modeled from a perspective view of supervised learning, roughly falling into the fields of information retrieval~\cite{burges2005learning,cao2007learning} and recommendation~\cite{linden2003amazon,koren2015advances}. For these methods, a Click-Through-Rate (CTR) model is usually built based on either a ranking function or a collaborative filtering system, then impressions are allocated according to the CTR scores. However, these methods usually optimize the short-term clicks, by assuming that the properties of products is independent of the decisions of the platform, which may hardly hold in the real E-commerce environment. There are also a few work trying to apply the mechanism design to the allocation problem from an economic theory point of view such as~\cite{myerson1981optimal,nisan2001algorithmic,shoham2008multiagent}. Nevertheless, these methods only work in very limited cases, such as the participants play only once, and their properties is statistically known or does not change over time, etc., making them far from practical use in our scenario. A recent pioneer work named \emph{Reinforcement Mechanism Design}~\cite{tang2017reinforcement} attempts to get rid of nonrealistic modeling assumptions of the classic economic theory and to make automated optimization possible, by incorporating the Reinforcement Learning (RL) techniques. It is a general framework which models the resource allocation problem over a sequence of rounds as a Markov decision process (MDP)~\cite{papadimitriou1987complexity}, and solves the MDP with the state-of-the-art RL methods. However, by defining the impression allocation over products as the action, it can hardly scale with the number of products/sellers as shown in~\cite{cai2018reinforcement,Cai2017ReinforcementMD}. Besides, it depends on an accurate behavioral model for the products/sellers, which is also unfeasible due to the uncertainty of the real world.


Although the properties of products can not be fully observed or accurately predicted, they do share a similar pattern in terms of  development trend, as summarized in the \emph{product lifecycle theory}~\cite{levitt1965exploit,cao2012product}. The life story of most products is a history of their passing through certain recognizable stages including  \emph{introduction},  \emph{growth},  \emph{maturity} and  \emph{decline} stages.

\begin{itemize}
\item{\emph{Introduction}}:   Also known as \emph{market development} - this is when a new product is first brought to market. Sales are low and creep along slowly.
\item{\emph{Growth}}: Demand begins to accelerate and the size of the total market expands rapidly.
\item{\emph{Maturaty}}: Demand levels off and grows.
\item{\emph{Decline}}: The product begins to lose consumer appeal and sales drift downward.
\end{itemize}
During the lifecycle, new products arrive continuously and outdated products wither away every day, leading to a natural metabolism in the E-commerce platform. Due to the insufficient statistics, new products usually attract few attention from conventional supervised learning methods, making the metabolism a very long period.


Inspired by the product lifecycle theory as well the reinforcement mechanism design framework, we consider to develop reinforcement mechanism design while taking advantage of the product lifecycle theory. The key insight is, with the power of trial-and-error, it is possible to recognize in advance the potentially hot products in the introduction stage as well as the potentially slow-selling products in the decline stage, so the metabolism can be speeded up and the long-term efficiency can be increased with an optimal impression allocation strategy.

We formally establish the lifecycle model and formulate the impression allocation problem by regarding the global status of products as the state and the parameter adjustment of a scoring function as the action. Besides, we develop a novel framework which incorporates a first principal component based algorithm and a repeated sampling based experiences generation method, as well as a shared convolutional neural network to further enhance the expressiveness and robustness. Moreover, we compare the feasibility and efficiency between baselines and the improved algorithms in a simulated environment built based on one of the largest E-commerce platforms.
 

%

The rest of the paper is organized as follows.  
The product lifecycle model and reinforcement learning algorithms are introduced in section~\ref{sec_preliminaries}. Then a reinforcement learning mechanism design framework is proposed in section~\ref{sec_rl}. Further more, experimental results are analyzed in section~\ref{sec_experiment}. Finally, conclusions and future work are discussed in section~\ref{sec_conclusion}.

\section{Related Work}
\label{sec_related_work}
Many researches have been conducted on impression allocation and dominated by supervised learning. In ranking phase, search engine aims to find out good candidates and brought them in front so that products with better performance will gain more impressions. Among which click-through rate is one of the most common representation of products performance. Some research presents an approach to automatically optimize the retrieval quality with well-founded retrieval functions under risk minimization frame-work by historical click-through data~\cite{joachims2002optimizing}. Some other research proposed an unbiased estimation of document relevance by estimating the presentation probability of each document~\cite{dupret2008user}. Nevertheless, both of these research suffer from low accuracy of click-through rate estimation for the lack of exposure historical data of start-ups.

One of the most related topics in user impressions allocation is \emph{item cold-start problem}~\cite{ricci2011introduction}, which has been extensively studied over past decades. Researches can be classified into three categories: hybrid algorithms combining CF with content-based techniques~\cite{park2009pairwise, saveski2014item},  bandit algorithms~\cite{liu2014promoting,anava2015budget,aharon2015excuseme} and data supplement algorithms~\cite{volkovs2017dropoutnet}. Among these researches, the hybrid algorithms exploit items' properties, the bandit algorithms are designed for no item  content setting and gathering interactions from user effectively, and the  data supplement algorithms view cold-start as data missing problem. 
Both of these research did not take the whole product lifecycle of items into account for the weakness of traditional prediction based machine learning model, {resulting in long-term imbalance between global efficiency and lifecycle optimization}. 

{The application of reinforcement learning in commercial system such as web recommendations and e-commerce search engines has not yet been well developed}. Some attempts are made to model the user impression allocation problem in e-commerce platform such as Tabao.com and Amazon.com. By regarding the platforms with millions of users as environment and treating the engines allocating user impressions as agents, an Markov Decision Process or at least Partially Observable Markov Decision Process can be established. For example, an reinforcement learning capable model is established on each page status by limit the page visit sequences to a constant number in a recommendation scene ~\cite{taghipour2007usage}. And another proposed model is established on global status by combining all the item historical representations in platform ~\cite{cai2018reinforcement}. However, both of these approaches struggled to manage an fixed dimensionality of state observation, low-dimensional action outputs and suffered from partially observation issues.

Recently, mechanism design has been applied in impression allocation, providing a new approach for better allocating user impressions~\cite{cai2016mechanism,tang2017reinforcement}. However, the former researches are not suitable for real-world scenes because of the output action space is too large to be practical.  
In this paper, a reinforcement learning based mechanism design is established for the impression allocation problem to maximize both short-term as well as long-term return of products in the platform with a new approach to extract states from all products and to reduce action space into practical level. 
\section{Preliminaries}
\label{sec_preliminaries}


\subsection{Product Lifecycle Model}

\label{sec_plc}

In this subsection, we establish a mathematical model of product lifecycle with noises. At step $t$,  each product has an observable attribute vector $x_t \in \mathbb{R}^d$ and an unobservable latent lifecycle state $z_t \in \mathcal{L}$, where $d$ is the dimension of the attribute space, and $\mathcal{L}=\{0,1,2,3\}$ is the set of lifecycle stages indicating the the \emph{introduction},  \emph{growth},  \emph{maturity} and  \emph{decline} stages respectively. Let $p_t \in \mathbb{R}$ be the CTR  and $q_t \in \mathbb{R}$ be the accumulated user impressions of the product. Without loss of generality, we assume $p_t$ and $q_t$ are observable, $p_t,q_t$ are two observable components of $x_t$, the platform allocates the impressions $u_t \in \mathbb{R}$ to the product. The dynamics of the system can be written as
\begin{align}
\left\{
\begin{aligned}
&q_{t+1} = q_t +u_t \\
&p_{t+1} = p_t +f(z_t,q_t) \\
&z_{t+1} = g(x_t,z_t,t)
\end{aligned}
\right.
\end{align}
where $f$ can be seen as the derivative of the $p$, and $g$ is the state transition function over $\mathcal{L}$.

According to the product lifecycle theory and online statistics, the derivative of the CTR can be formulated as 
\begin{align}
f(z_t,q_t) = \left\{
\begin{aligned}
& \frac{(c_h-c_l)e^{-\delta(q_t) }}{{(2-z)(1+e^{-\delta(q_t)})^2}} + \xi, &z\in\{1, 3\} \\
&\xi, & z\in\{0, 2\} \\
\end{aligned}
\right.
\end{align}
where $\xi \sim  \mathcal{N}(0,\sigma^2)$ is a gaussian noise with zero mean and variance $\sigma^2$, $\delta(q_t)=(q_t-\tilde{q_t}_z - \delta_\mu)/\delta_\sigma$ is the normalized impressions accumulated from stage $z$ ,  $\tilde{q_t}_z$ is the initial impressions when the product is firstly evolved to the life stage $z$, $\delta_\mu,\delta_\sigma$ are two unobservable parameters for normalization, and $c_h,c_l \in \mathbb{R}$ are the highest CTR and the lowest CTR during whole product lifecycle, inferred from two neural networks, respectively:
\begin{align}
c_l = h(x_t|\theta_l),\;\; c_h = h(x_t|\theta_h),
\end{align}
where $h(\cdot|\theta)$ is a neural network with the fixed parameter $\theta$, indicating that $c_l,c_h$ are unobservable but relevant to attribute vector $x_t$. Intuitively, when the product stays in introduction or maturity stage, the CTR can be only influenced by the noise. When the product in the growth stage, $f$ will be a positive increment, making the CTR increased up to the upper bound $c_h$. Similar analysis can be obtained for the product in the decline stage. 
\begin{figure}[!h]
\centering\includegraphics[width=0.8\linewidth]{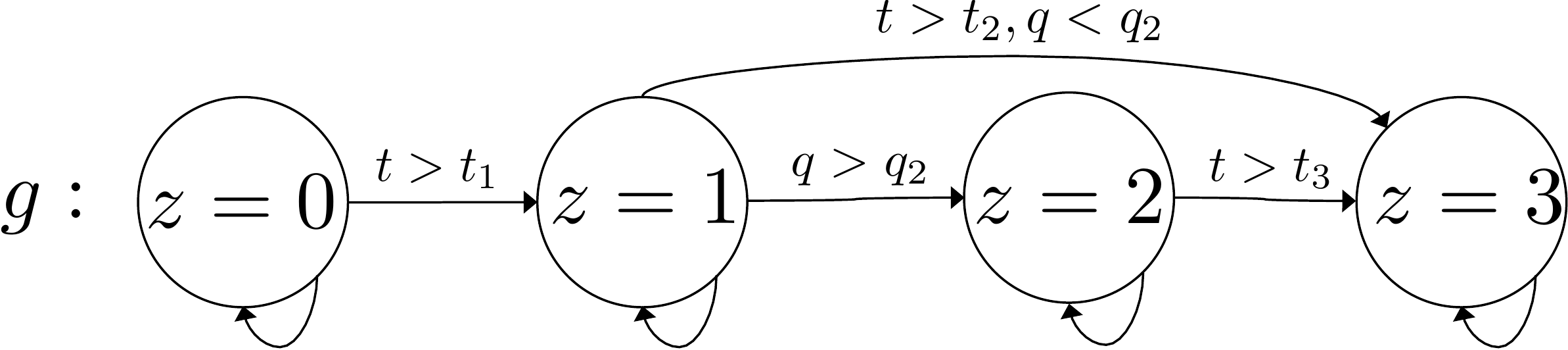}
\caption{State transition during product lifecycle}
\label{fig:state_trans}
\end{figure}

Then we define the state transition function of product lifecycle as a finite state machine as illustrated in Fig. \ref{fig:state_trans}. The product starts with the initial stage $z=0$, and enters the growth stage when the time exceeds $t_1$. During the growth stage, a product can either step in to the maturity stage if its accumulated impressions $q$ reaches $q_2$, or the decline stage if the time exceeds $t_2$ while $q$ is less than $q_2$. A product in the maturity stage will finally enter the last decline stage if the time exceeds $t_3$. Otherwise, the product will stay in current stage. Here, $t_1,t_2,t_3,q_2$ are the latent thresholds of products. 

\begin{figure}[t]
\includegraphics[width=1\linewidth]{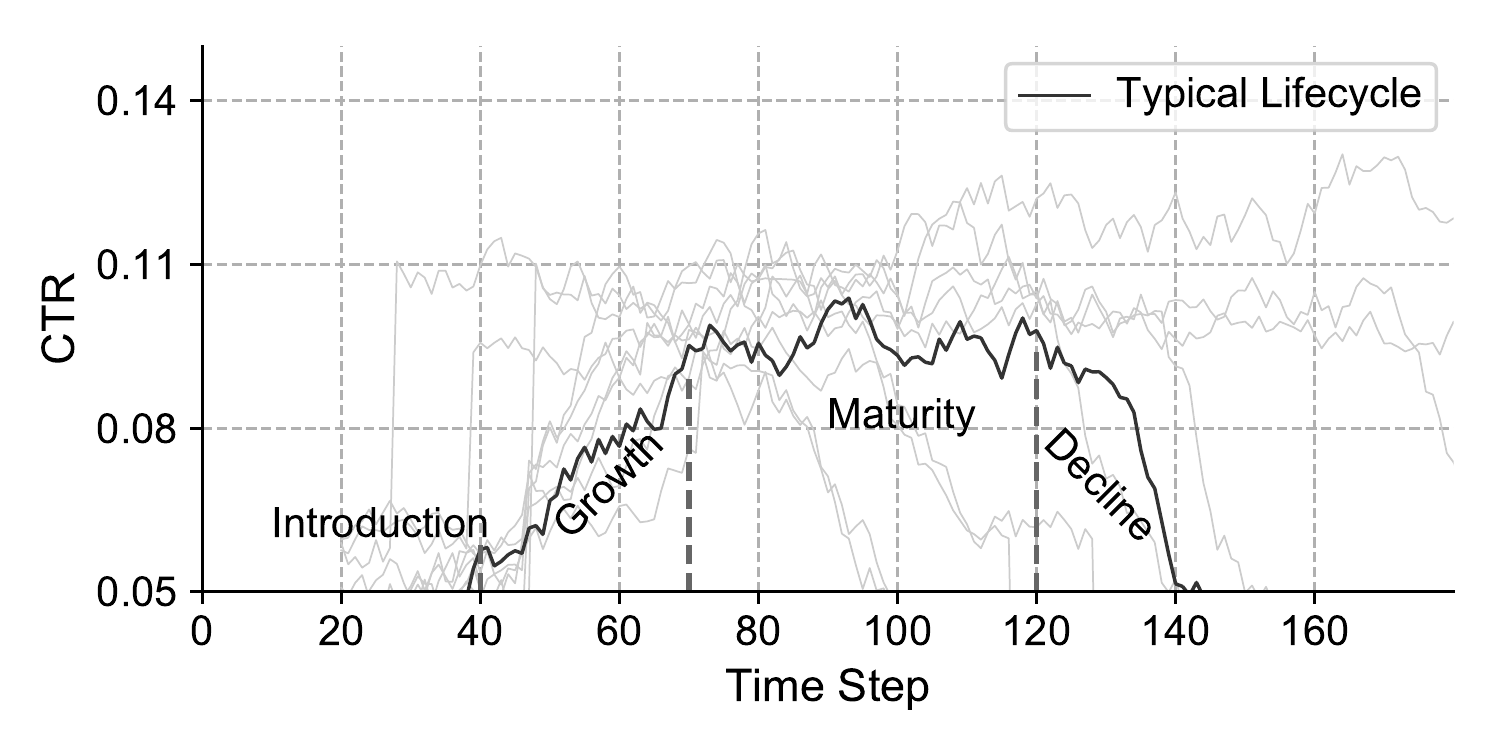}
\caption{CTR evolution with the proposed lifecycle model.}
\label{fig:plc}
\end{figure}

We simulate several product during the whole lifecycle with different latent parameters (the details can be found in the experimental settings), the CTR curves follow the exact trend described in Fig. \ref{fig:plc}.

\subsection{Reinforcement Learning and DDPG methods}
Reinforcement learning maximizes accumulated rewards by trial-and-error approach in a sequential decision problem. The sequential decision problem is formulated by MDP as a tuple of state space $\mathcal{S}$, action space $\mathcal{A}$, a conditional probability distribution $p(\cdot)$ and a scalar reward function $r = R({s},{a}), R:\mathcal{S} \times \mathcal{A} \rightarrow \mathbb{R}$. For states ${s}_t, {s}_{t+1} \in \mathcal{S}$ and action ${a}_t \in \mathcal{A}$, distribution function $p({s}_{t+1}|{s}_t, {a}_t)$ denotes the transition probability from state ${s}_t$ to ${s}_{t+1}$ when action ${a}_t$ is adopted in time step $t$, and the Markov property $p({s}_{t+1}|{s}_t, {a}_t) = p({s}_{t+1}|{s}_1, {a}_1, \cdots, {s}_t, {a}_t)$ holds for any historical trajectories 
${s}_1, {a}_1, \cdots, {s}_t$ to arrive at status ${s}_t$. A future discounted return at time step $t$ is defined as $R_t^{\gamma}=\sum_{k=t}^{\infty}\gamma^{k-t}R({s}_k, {a}_k)$, where $\gamma$ is a scalar factor representing the discount.  A policy is denoted as $\pi_{{\theta}}({a}_t | {s}_t)$ which is a probability distribution mapping from $\mathcal{S}$ to $\mathcal{A}$ , where different policies are distinguished by parameter ${\theta}$. 

The target of agent in reinforcement learning is to maximize the expected discounted return, and the performance objective can be denoted as
\begin{align}
\max \limits_\pi J &=\mathbb{E}\left[R_1^{\gamma}\right|\pi] \notag\\
&=  \mathbb{E}_{{s} \sim d^{\pi}, {a} \sim \pi_{{\theta}}}\left[R({s}, {a}) \right]
\end{align}
where $d^{\pi}({s})$ is a discounted state distribution indicating the possibility to encounter a state ${s}$ under the policy of $\pi$. An action-value function is then obtained iteratively as
\begin{align}
Q({s}_t, {a}_t) = \mathbb{E}\left[R({s}_{t}, {a}_{t}) + \gamma\mathbb{E}_{ {a} \sim \pi_{{\theta}}}\left[Q({s}_{t+1}, {a}_{t+1})\right]\right]
\end{align}
In order to avoid calculating the gradients of the changing state distribution in continuous action space, the Deterministic Policy Gradient(DPG) method~\cite{sutton2000policy,silver2014deterministic} and the Deep Deterministic Policy Gradient~\cite{lillicrap2015continuous} are brought forward. Gradients of the deterministic policy $\pi$ is
\begin{align}
\nabla _{{\theta}^{\mu}} J &= \mathbb{E}_{{s}\sim d^{\mu}} \left[ \nabla _{{\theta}^{\mu}}Q^{{w}}({s}, {a})\right] \notag\\
&= \mathbb{E}_{{s} \sim d^{\mu}}\left[\nabla_{{\theta}^{\mu}} \mu({s})\nabla_{{a}}Q^{{w}}({s}, {a}) |_{{a}={\mu}({s})}\right]
\end{align}
where $\mu$ is the deep actor network to approximate policy function. And the parameters of actor network can be updated as
\begin{align}
{\theta}^{\mu} \leftarrow {\theta}^{\mu} + \alpha \mathbb{E}\left[\nabla_{{\theta}^{\mu}} \mu({s_t})\nabla_{{a}}Q^{{w}}({s_t}, {a_t}) |_{{a}={\mu}({s})}\right]
\end{align}
where $Q^{{w}}$ is an obtained approximation of action-value function called critic network. Its parameter vector ${w}$ is updated according to objective
\begin{equation}
\min \limits _{{w}} L = \mathbb{E}_{s\sim d^{\mu}}\left[y_t - Q^{{w}}({s_t}, {a_t}))^2\right]
\label{critic_loss}
\end{equation}
where $y_t=R({s_t}, {a_t}) + \gamma Q^{w'}({s_{t+1}}, \mu{'}({s_{t+1}}))$, ${\mu}'$ is the target actor network to approximate policy ${\pi}$,  $Q^{{w}'}$ is the target critic network to approximate action-value function. The parameters $w{'}, \theta^{\mu{'}}$ are updated softly as
\begin{align}
{w}' &\leftarrow \tau{w}' + (1-\tau ) {w}\notag\\
{\theta}^{{\mu}'}&\leftarrow\tau{\theta}^{{\mu}'}+ (1 - \tau) {\theta}^{{\mu}}
\end{align}

\section{A Scalable Reinforcement Mechanism Design Framework}
\label{sec_rl}
In our scenario, at each step, the platform observes the global information of all the products, and then allocates impressions according to the observation and some certain strategy, after which the products get their impressions and update itself with the attributes as well as the lifecycle stages. Then the platform is able to get a feedback to judge how good its action is, and adjust its strategy based on all the feedbacks. The above procedures leads to a standard sequential decision making problem. 

However, application of reinforcement learning to this problem encounters sever computational issues, due to high dimensionality of  both action space and state space, especially with a large $n$. Thus, we model the impression allocation problem as a standard reinforcement learning problem formally, by regarding the global information of the platform as the state
\begin{align}
s = [x_1,x_2,...,x_n]^{\rm{T}} \in \mathbb{R}^{n \times d}
\end{align}
where $n$ is the number of the product in the platform, $d$ is the dimension of the attribute space, and regarding the parameter adjustion of a score function as the action,
\begin{align}
a = \pi(s|\theta^{\mu}) \in \mathbb{R}^d
\end{align}
where $\pi$ is the policy to learn parameterize by $\theta^{\mu}$, and the action $a$ can be further used to calculate scores of all products
\begin{align}
o_i =  \frac{1}{1+e^{-a^{\rm{T}}x_i}},\;\; \forall i \in \{1,2,...,n\}
\end{align}
After which the result of impression allocation over all $n$ products can be obtained by
\begin{align}
u_i =  \frac{e^{o_i}}{\sum_i^n e^{o_i}},\;\; \forall i \in \{1,2,...,n\}
\end{align}
Without loss of generosity, we assume at each step the summation of impressions allocated is 1, i.e., $\sum_i^n u_i =1$. As is well known, products number $n$(billions) is far bigger than products attributes dimensions $d$(thousands) in large scale E-commerce platforms. By such definition, the dimension of the action space is reduced to $d$, significantly alleviating the computational issue in previous work~\cite{Cai2017ReinforcementMD}, where the the dimension of the action space is $n$. 

The goal of policy is to speeded up the metabolism by scoring and ranking products under the consideration of product lifecycle, making the new products grow into maturity stage as quickly as possible and keeping the the global efficiency from dropping down during a long term period. Thus, we define the reward related to $s$ and $a$ as 
\begin{equation}
R(s,a)=\frac{1}{n} \sum_i^n \left[\frac{1}{t_i}{\int\limits_{t=0}^{t_i} p(t) \frac{d{q(t)}}{dt}dt}\right]
\label{equ_rewards}
\end{equation}
where $t_i$ is the time step of the $i$-th product after being brought to the platform, $p(t), q(t)$ is the click through rate function and accumulated impressions of a product respectively. The physical meaning of this formulation is the mathematical expect over all products in platform for the average click amount of an product during its lifecycle, indicating the efficiency of products in the platform and it can be calculated accumulatively in the online environment, which can be approximately obtained by 
\begin{equation}
R(s,a)\approx \frac{1}{n} \sum_i^n \frac{1}{t_i} \sum_{\tau=0}^{t_i}  p_{\tau}^i u_{\tau}^i
\end{equation}

A major issue in the above model is that, in practices there will be millions or even billions of products, making combinations of all attribute vectors to form a complete system state with size $n \times d$ computationally unaffordable as referred in essays ~\cite{cai2018reinforcement}. A straightforward solution is to applying feature engineering technique to generate a low dimension representation of the state as ${s}_l = \mathcal{G}({s})$, where $\mathcal{G}$ is a pre-designed aggregator function to generate a low dimensional representation of the status. However, the pre-designed aggregator function is a completely subjective and highly depends on the the hand-craft features. Alternatively,  we attempt to tackle this problem using a simple sampling based method. Specifically, the state is approximated by $n_s$ products uniformly sampled from all products
\begin{equation}
\hat{s} = [{x}_1, {x}_2, \cdots, {x}_{n_s}]^{\rm{T}} \in \mathbb{R}^{n_s \times d}
\end{equation}
where $\hat{s}$ is the approximated state. Then, two issues arise with such sampling method:
\begin{itemize}
\item{In which order should the sampled ${n_s}$ products  permutated in $\hat{s}$, to implement the \emph{permutation invariance}?}
\item{How to reduce the bias brought by the sampling procedure, especially when $n_s$ is much smaller than $n$?}
\end{itemize}
To solve these two problem, we further propose the first principal component based permutation and the repeated sampling based experiences generation, which are described in the following subsections in details.

\subsection{First Principal Component based Permutation}
\label{sec_sampling_based_state_observation}
The order of each sampled product in the state vector has to be proper arranged, since the unsorted state matrix vibrates severely during training process, making the parameters in network hard to converge. To avoid it, a simple way for permutation is to make order according to a single dimension, such as the brought time $t_i$, or the accumulated impressions $q_i$. However, such ad-hoc method may lose information due to the lack of general principles. For example, if we sort according to a feature that is almost the same among all products, state matrix will keep vibrating severely between observations. A suitable solution is to sort the products in an order that keep most information of all features, where the first principal components are introduced ~\cite{abdi2010principal}. We design a first principal component based permutation algorithm, to project each $x_i$ into a scalar $v_i$ and sort all the products according to $v_i$
\begin{align}
&{e}_t=\arg\max\limits_{\lVert {e}\rVert=1} \left({e}^{\rm{T}}{s_t}^{\rm{T}} {s_t}{e}\right)\label{eq:pca}\\
&\hat{{e}} = \frac{\beta \hat{{e}} + (1-\beta)\left({e}_t - \hat{e}\right)}{\lVert \beta \hat{{e}} + (1-\beta)\left({e}_t - \hat{{e}}\right)\rVert} \label{eq:softupdate}\\
&v_i = \hat{{e}}^{\rm{T}}x_i , i = 1, 2, \cdots, n_s \label{eq:proj}
\end{align}
where ${e}_t$ is the first principal component of system states in current step $t$ obtained by the classic PCA method as in Eq. \ref{eq:pca}.  $\hat{{e}}$ is the projection vector softly updated by ${e}_t$ in Eq. \ref{eq:softupdate}, with which we calculate the projected score of each products in Eq. \ref{eq:proj}. Here $0<\beta<1$ is a scalar indicating the decay rate of $\hat{e}$.  Finally, the state vector is denoted as 
\begin{equation}
\hat{s}= [{x}_{k_1}, {x}_{k_2}, \cdots, {x}_{k_{n_s}}]^{\rm{T}}
\end{equation}
where $k_1, k_2, \cdots, k_{n_s}$ is the order of products, sorted by $v_i$. 

\subsection{Repeated Sampling based Experiences Generation}
We adopt the classic experience replay technique~\cite{Lin1992Self,mnih2013playing} to enrich experiences during the training phase just as other reinforcement learning applications. In the traditional experience replay technique, the experience is formulated as $({s}_t, {a}_t, r_t, {s}_{t+1})$. However, as what we describe above, there are $\mathrm{C}_n^{n_s}$ observations each step theoretically, since we need to sample $n_s$ products from all the $n$ products to approximate the global statistics. If $n_s$ is much smaller than $n$, such approximation will be inaccurate. 

\begin{figure}[h]
\centering\includegraphics[width=1\linewidth]{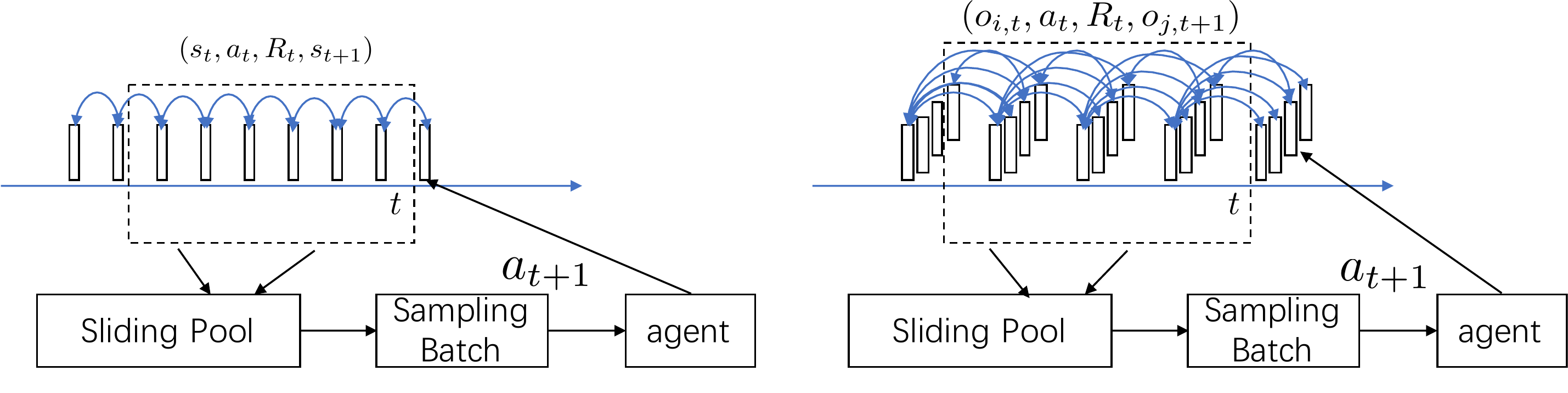}
\caption{Classical experiences generation(left): One experience is obtained each step by pair$({s}_t, {a}_t, r_t, {s}_{t+1})$; Repeated sampling based experiences generation(right): $m^2$ experiences are obtained each step by pair$(\hat{s}_t^i, {a}_t, r_t, \hat{s}_{t+1}^j)$ }
\label{fig:experience}
\end{figure}

To reduce the above bias, we propose the repeated sampling based experiences generation. For each original experience, we do repeated sampling ${s}_t$ and ${s}_{t+1}$ for $m$ times, to obtain $m^2$ experiences of
\begin{equation}
(\hat{s}_t^i, {a}_t, r_t, \hat{s}_{t+1}^j), \;\; i,j \in 1, 2,\cdots,m
\end{equation}
as illustrated in Fig.~\ref{fig:experience}. 
This approach improves the stability of observation in noise environment. It is also helpful to generate plenty of experiences in the situation that millions of times repetition is unavailable. 

It is worth noting that, the repeated sampling is conducted in the training phase. When to play in the environment, the action $a_t$ is obtained through a randomly selected approximated state $\hat{s}_t$, i.e., $a_t=\pi(\hat{s}_t^1)$. Actually, since $a_t$ does not necessarily equal to $\pi(\hat{s}_t^i),  \forall i \in 1,2, \cdots, m$, it can further help learning a invariant presentation of the approximated state observations.

\begin{algorithm}[!t]
  \SetAlgoNoLine
  \LinesNumbered
  Initialize the parameters of the actor-critic network ${\theta}^{\mu}, {w}, {\theta^{\mu{'}}}, {w{'}}$\\
  Initialize the replay buffer $M$\\
  Initialize $m$ observations $\hat{s}_{0}^{j}$\\
  Initialize the first principal component  $\hat{p}$ by $\hat{s}_0$\\
  \ForEach{training step $t$}{
      Select action ${a}_t = \mu({\hat{s}}_{t}^1|\theta^{\mu})$\\
      Execute action ${a}_t$ and observe reward ${r}_t$\\
      \ForEach{$j \in 1,2, \cdots, m$}{
          Sample a random subset of $n_s$ products\\
          Combine an observation in the order of ${x}_k^{\rm{T}}\hat{e}$\\
          \vspace{-3mm}$${\hat{s}}_{t}^j \leftarrow  \left({x}_{k_1}, {x}_{k_2}, \cdots, {x}_{k_{n_s}}\right)^{\rm{T}}$$
          Update first principal component \\
          \vspace{-3mm}$${e_t}\leftarrow \arg\max_{\lVert {e}\rVert=1} \left({e}^{\rm{T}}{{\hat{s}_{t}^{j\rm{T}}}}{ {\hat{s}}_{t}^j}{e}\right)$$
          \vspace{-3mm}$$\hat{e} \leftarrow \textnormal{norm}\left(\beta \hat{e} + (1-\beta)\left({e_t} - \hat{e}\right)\right)$$

      }
      
      \ForEach{$i,j \in 1,2,\cdots, m$}{  $$M \leftarrow M \cup \{ (\hat{s}_t^i, {a}_t, r_t, \hat{s}_{t+1}^j)\}$$\\
      }
      
      Sample $n_k$ transitions from ${{M}}$: $({\hat{s}}_{k}, {a}_{k}, {r}_k, {\hat{s}}_{k+1})$ \\
      Update critic and actor networks
      \begin{align}
       \vspace{-3mm}&w\leftarrow w + \frac{\alpha_w }{n_k}\sum_k( y_k - Q^{{w}}({\hat{s}}_k, {a}_k))\nabla_{w}Q^{{w}}({\hat{s}}_k, {a}_k)\notag\\
      \vspace{-5mm}&{\theta}^{\mu} \leftarrow {\theta}^{\mu} + \frac{\alpha_{\mu}}{n_k}\sum_{k}\nabla_{{\theta}^{\mu}} \mu({{\hat{s}}_k})\nabla_{{a}_k}Q^{{w}}({\hat{s}}_k, {a}_k) \notag
      \end{align}
      Update the target networks\\
      \vspace{-3mm}$${w}' \leftarrow \tau{w}' + (1-\tau ) {w}$$
      \vspace{-3mm}$${\theta}^{{\mu}'} \leftarrow \tau{\theta}^{{\mu}'}+ (1 - \tau) {\theta}^{{\mu}}$$

  }
  \caption{The Scalable Reinforcement Mechanism Design Framework}
  \label{alg:paper}
\end{algorithm} 

The overall procedure of the algorithm is described in Algorithm~\ref{alg:paper}. Firstly, a random sampling is utilized to get a sample of system states.  And then the sample is permutated by the projection of the first principal components. After that, a one step action and multiple observations are introduced to enrich experiences in experience pool. Moreover, a shared convolutional neural network is applied within the actor-critic networks and target actor-critic networks to extract features from the ordered state observation~\cite{cheng2004application,wu2016training}. Finally, the agent observes system repeatedly and train the actor-critic network to learn an optimized policy gradually. 

\section{Experimental Results}
\label{sec_experiment}

To demonstrate how the proposed approach can help improve the long-term efficiency by speeding up the metabolism, we apply the proposed reinforcement learning based mechanism design, as well as other comparison methods, to a simulated E-commerce platform built based on the proposed product lifecycle model. 

\subsection{The Configuration}
The simulation is built up based on product lifecycle model proposed in section~\ref{sec_plc}. Among all of the parameters, $q_2$ is uniformly sampled from $[10^4, 10^6]$, $t_1, t_2, t_3, \delta_\mu, \delta_\sigma$ are uniformly sampled from $[5, 30], [35, 120], [60, 180], [10^4, 10^6], [2.5\times10^3, 2.5\times10^5]$ respectively, and parameter $\sigma$ is set as $0.016$ . The parameters $c_l, c_h$ are generated by a fixed neural network whose parameter is uniformly sampled from $[-0.5, 0.5]$ to model online environments, with the outputs scaled into the intervals of $[0.01, 0.05]$ and $[0.1, 0.15]$ respectively. Apart from the normalized dynamic CTR $p$ and the accumulated impressions $q$, the attribute vector $x$ is uniformly sampled from $[0,1]$ element-wisely with the dimension $d=15$. All the latent parameters in the lifecycle model are assumed unobservable during the learning phase.

The DDPG algorithm is adopted as the learning algorithm. The learning rates for the actor network and the critic network are $\rm{10}^{-4}$ and  $\rm{10}^{-3}$ respectively, with the optimizer ADAM~\cite{journalscorrKingmaB14}. The replay buffer is limit by $2.5\times10^{4}$. The most relevant parameters evolved in the learning procedure are set as table~\ref{tab:param}.
\begin{table}
  \center\caption{Parameters in learning phase}
  \label{tab:param}
  \begin{tabular}{ccl}
    \toprule
    Param & Value & Reference\\
    \midrule
    $n_s$ & $10^{3}$& Number of products in each sample \\
    $\beta$& $0.999$& First principal component decay rate \\
    $\gamma$ & $0.99$ & Rewards discount factor\\
    $\tau$ & $0.99$ & Target network decay rate\\
    $m$ & $5$ & Repeated observation times\\
    \bottomrule
  \end{tabular}
\end{table}

Comparisons are made within the proposed reinforcement learning based methods as
\begin{itemize}
\item{\textbf{CTR-A}: The impressions are allocated in proportion to the CTR score. }
\item{\textbf{T-Perm}: The basic DDPG algorithm, with brought time based permutation and a fully connected network to process the state}
\item{\textbf{FPC}: The basic DDPG algorithm, with first principal component based permutation and a fully connected network to process the state. }
\item{\textbf{FPC-CNN}:  {FPC} with a shared two-layers convolutional neural network in actor-critic networks. }
\item{\textbf{FPC-CNN-EXP}: {FPC-CNN} with the improved experiences generation method. }
\end{itemize}
where {CTR-A} is the classic supervised learning method and the others are the proposed methods in this paper. For all the experiments, {CTR-A} is firstly applied for the first 360 steps to initialize system into a stable status, i.e., the distribution over different lifecycle stages are stable, then other methods are engaged to run for another $2\rm{k}$ steps and the actor-critic networks are trained for $12.8\rm{k}$ times.

\subsection{The Results}

\begin{figure}[t]
\includegraphics[width=1\linewidth]{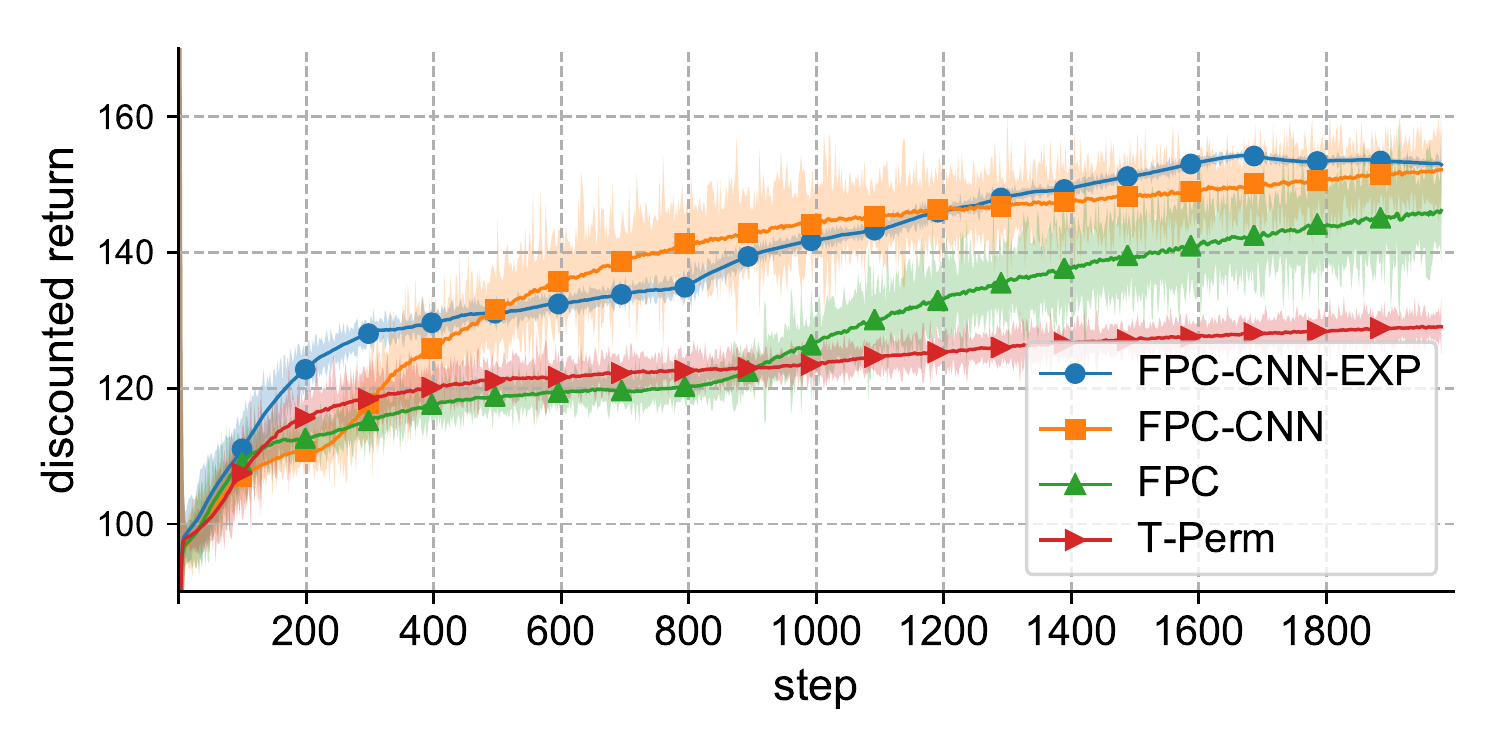}
\caption{Performance Comparison between algorithms}
\label{fig:rewards}
\end{figure}
We firstly show the discounted accumulated rewards of different methods at every step in Fig.  ~\ref{fig:rewards}. After the initialization with the CTR-A, we find that the discounted accumulated reward of CTR-A itself almost converges to almost 100 after 360 steps (actually that why 360 steps is selected for the initialization), while that of other methods can further increase with more learning steps. It is showed that all FPC based algorithms beat the T-Perm algorithm, indicating that the FPC based algorithm can find a more proper permutation to arrange items while the brought time based permutation leads to a loss of information, making a drop of the final accumulated rewards. Moreover, CNN and EXP algorithms perform better in extracting feature from observations automatically, causing a slightly improvement in speeding up the converging process. Both the three FCP based algorithms converge to same final accumulated rewards for their state inputs have the same observation representation. 

\begin{figure}[h]
\includegraphics[width=1\linewidth]{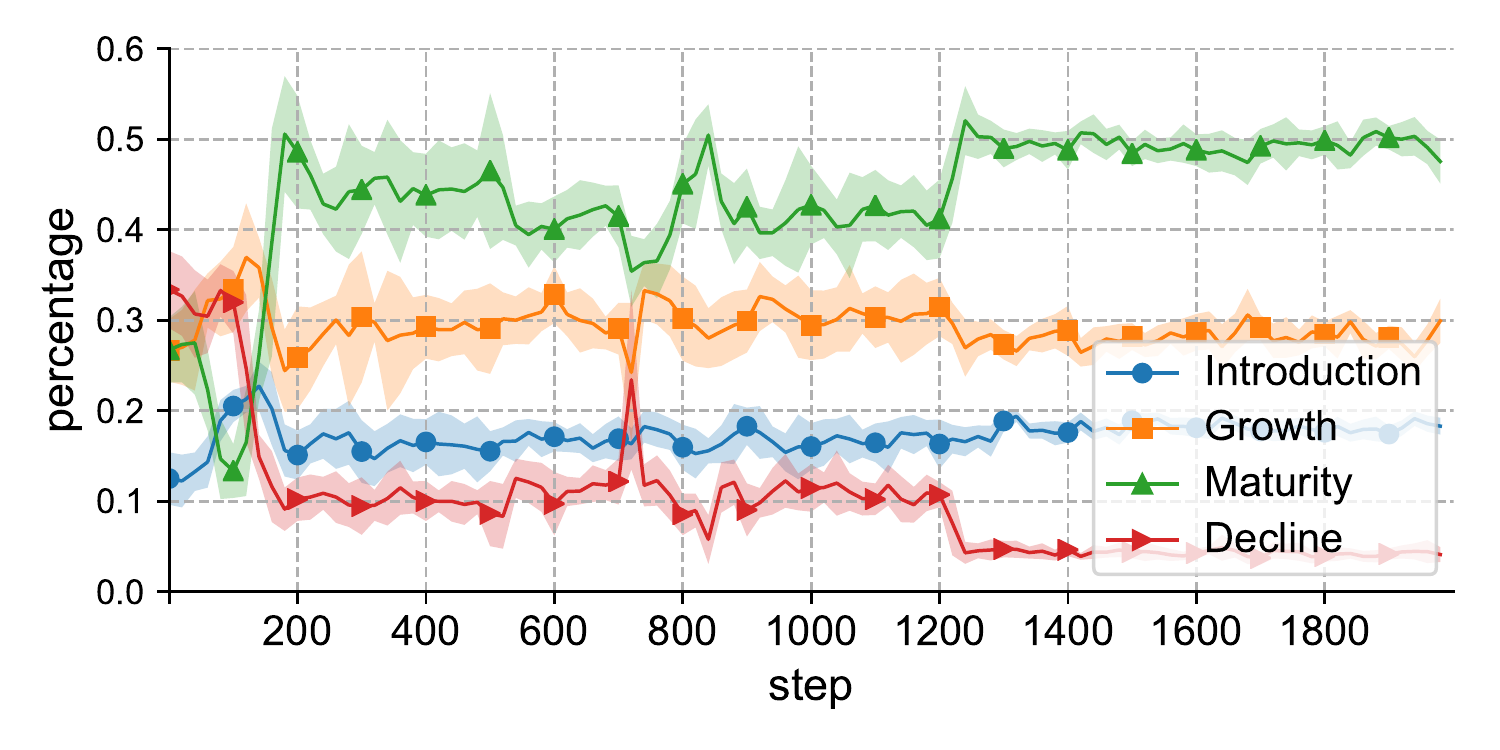}
\caption{Percentage of impressions allocated to different stages.}
\label{fig:dq_percentage}
\end{figure}

Then we investigate the distribution shift of the  impression allocation over the 4 lifecycle stages after  the training procedure of the FPC-CNN-EXP method, as shown in Fig.~\ref{fig:dq_percentage}. It can be seen that the percentage of decline stage is decreased and percentage of introduction and maturity stages are increased. By giving up the products in the decline stage, it helps the platform to avoid the waste of the impressions since these products are always with a low CTR. By encouraging the products in the introduction stage, it gives the changes of exploring more potential hot products. By supporting the products in the maturity stage, it maximizes the short-term efficiency since the they are with the almost highest CTRs during their lifecycle.

\begin{figure}[h]
\includegraphics[width=1\linewidth]{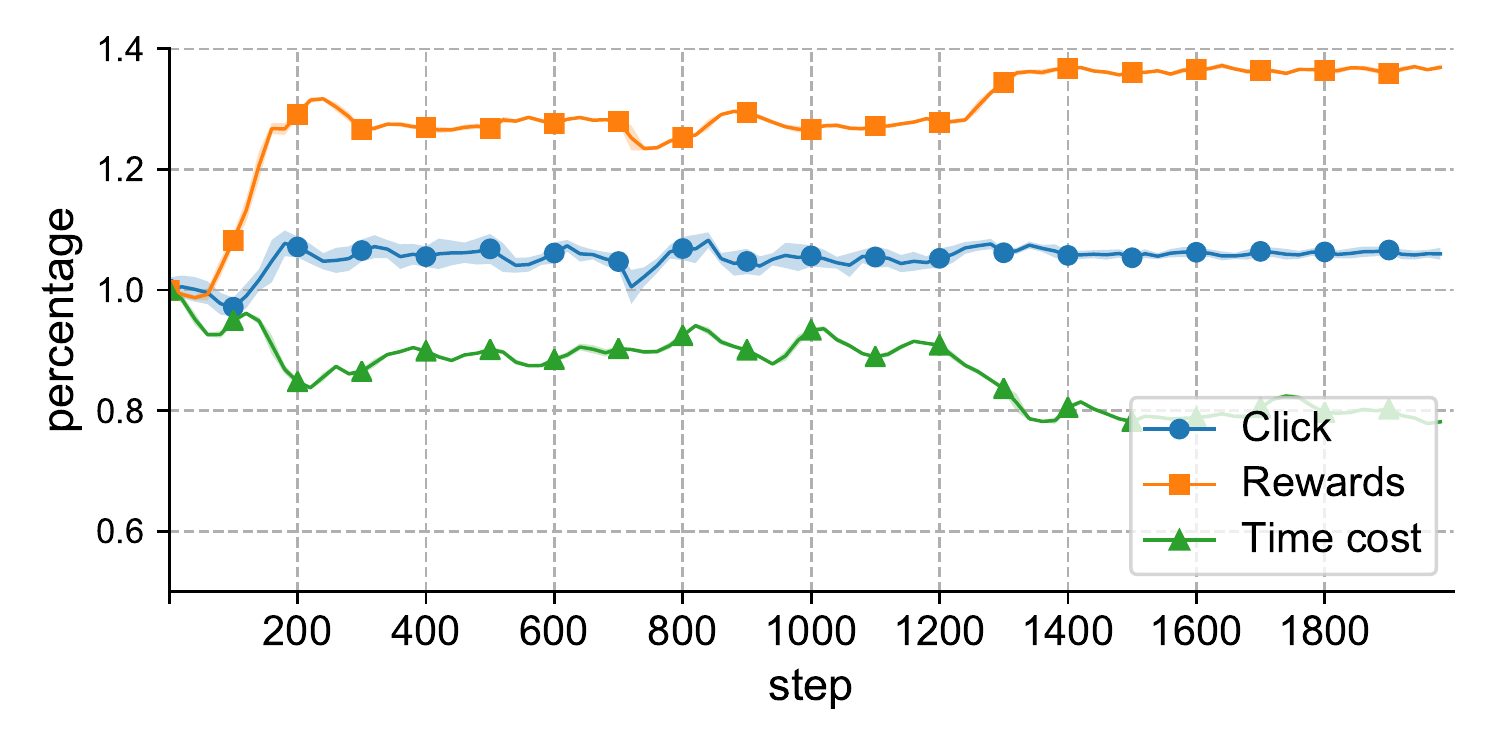}
\caption{Metabolism relative metrics}
\label{fig:metrics}
\end{figure}

We finally demonstrate the change of the global clicks, rewards as well as the averaged time durations for a product to grow up into maturity stage from its brought time at each step, in terms of relative change rate compared with the CTR-A method, as is shown in Fig. ~\ref{fig:metrics}. The global average click increases by 6\% when the rewards is improved by 30\%. The gap here is probably caused by the inconsistency of the reward definition and the global average click metric. In fact, the designed reward contains some other implicit objectives related to the metabolism. To further verify the guess, we show that the average time for items to growth into maturity stage has dropped by 26\%, indicating that the metabolism is significantly speeded up. Thus, we empirically prove that, through the proposed reinforcement learning based mechanism design which utilizes the lifecycle theory, the long-term efficiency can be increased by speeding up the metabolism.

\section{Conclusions and Future Work}
\label{sec_conclusion}
In this paper, we propose an end-to-end general reinforcement learning framework to 
improve the long-term efficiency by speeding up the metabolism. We reduce action space into a reasonable level and then propose a first principal component based permutation for better observation of environment state. After that, an improved experiences generation technique is engaged to enrich experience pool. Moreover, the actor-critic network is improved by a shared convolutional network for better state representation. Experiment results show that our algorithms outperform the baseline algorithms. 

For the future work, one of the promising directions is to develop a theoretical guarantee for first principal component based permutation. Another possible improvement is to introduce the  nonlinearity to the scoring function for products.


%


\appendix

\bibliographystyle{ACM-Reference-Format}
\bibliography{product_life_cycle_sigir}

\end{document}